\documentclass[pmlr,twocolumn,10pt]{jmlr} 

\usepackage{multirow}
\usepackage{booktabs}
\usepackage{threeparttable}

\usepackage{siunitx}

\theorembodyfont{\upshape}
\theoremheaderfont{\scshape}
\theorempostheader{:}
\theoremsep{\newline}

\title{A Quantitative and Qualitative Analysis of Schizophrenia Language}

\author{
\Name{Amal Alqahtani} \Email{amalqahtani@gwu.edu}\\
\addr The George Washington University, USA
\AND
\Name{Efsun Sarioglu Kayi} \Email{Efsun@gwu.edu}\\
\addr The George Washington University, USA
\AND
\Name{Sardar Hamidian} \Email{sardar@gwu.edu}\\
\addr The George Washington University, USA
\AND
\Name{Michael Compton} \Email{mtc2176@cumc.columbia.edu}\\
\addr The Columbia University, USA
\AND
\Name{Mona Diab} \Email{mdiab@fb.com}\\
\addr Meta AI, USA
}

\begin{document}

\maketitle

\begin{abstract}
Schizophrenia is one of the most disabling mental health conditions to live with. Approximately one percent of the population has schizophrenia which makes it fairly common, and it affects many people and their families. Patients with schizophrenia suffer different symptoms: formal thought disorder (FTD), delusions, and emotional flatness. In this paper, we quantitatively and qualitatively analyze the language of patients with schizophrenia measuring various linguistic features in two modalities: speech and written text. We examine the following features: coherence and cohesion of thoughts, emotions, specificity, level of committed belief (LCB), and personality traits. Our results show that patients with schizophrenia score high in fear and neuroticism compared to healthy controls. In addition, they are more committed to their beliefs, and their writing lacks details. They score lower in most of the linguistic features of cohesion with significant p-values.

\end{abstract}

\paragraph*{Data and Code Availability}

This paper uses the dataset of \cite{kayi2018predictive}. The dataset is not published publicly due to its sensitivity. We use different Python packages such as lm scorer \citep{Simone2020}, specificity \citep{gao2019specificity}, emotion \citep{abdul2017emonet} to generate the results, and these libraries are available online except coh-metrix \citep{graesser2004coh} where we grant the access by the author.

\section{Introduction}
\label{sec:intro}
Schizophrenia is a mental illness that can disrupt thought processes and perception \citep{kerns2002cognitive}. It can impair people's ability to manage their emotions, and can cause motor and behavioral disorders \citep{elvevag2000cognitive}.

Understanding and identifying the underlying signs of schizophrenia is critical in early detection and intervention before the malady becomes severely disabling if left untreated \citep{seeber2005does}. Moreover, it is vital to support mental health practitioners as well as policymakers to eliminate barriers to treating mental illnesses such as schizophrenia.

Gradual decline in functioning and cognition are some common characteristics of schizophrenia patients. Symptoms may include delusions, which are fixed false beliefs, as well as hallucinations but also importantly, they tend to have strong convictions regardless of the veridicality of the beliefs themselves. Another symptom that some individuals with schizophrenia exhibit is formal thought disorder (FTD), where a patient becomes unable to form coherent or logical thoughts \citep{kuperberg2010language}. Moreover, they suffer in some cases from lack of motivation and/or emotional response.

One way to capture mental disorders and related symptomatology is by analyzing patients' linguistic cues. Hence, we map the aforementioned symptoms to linguistic features that we can measure. To date, most of the employed measures used by clinicians measure superficial linguistic cues and they tend to be more qualitative. We hypothesize that advances in pragmatic NLP tools allow us to measure many of these symptoms via analyzing language cues used by patients. We surmise that given such tools, we help create objective quantitative measures for clinicians beyond what they are using today for diagnostics. Moreover having such tools could help them discover and codify further studies allowing for even more signals in detecting such mental health disorders.\footnote{Despite our focus in this work on schizophrenia, we believe that many of the tools we use here could be applicable to other mental disorders.} Accordingly, we present the first comprehensive study of deep pragmatically oriented linguistic modeling tools for diagnostic purposes. We leverage an emotion detection model to assess the lack of emotional response. We also employ a personality detection model to measure lack of motivation, which is one of the negative symptoms they may exhibit. We use a level of committed belief detection model to identify the level of committed belief corresponding to strength of conviction. Formal Thought Disorder (FTD) is measured by using language model-based sentence scoring as well as other coherence features such as LSA, connectives, lexical diversity, syntactic complexity, word information, and level of linguistic specificity. Finally, we employ the \emph{Coh-metrix} computational tool for analyzing texts for a variety of cohesion measures.
\citep{graesser2004coh}.

Accordingly, we investigate the following metrics: cohesion, level of committed belief, emotion, and personality and their corresponding correlation with symptomatic patients' language use. We examine both speech and text modalities, comparing patients vs. a matched set of controls.

Our results show that when patients express their emotions in writing or speech, they tend to show fear more often than other emotions. The findings also detect a neuroticism personality as they may suffer from feelings such as anger, and anxiety more frequently and severely. Furthermore, the results indicate that their writings lack specificity (details), and they are more committed to their beliefs in contrast with healthy controls. In addition, our results show that writings of healthy controls are more coherent demonstrated via the high scores of language model probabilities of their writing. To the best of our knowledge, our findings present the first set of measurable pragmatic linguistic cues that significantly correlate with contrastive mental health patients' language use that goes beyond the typical superficial metrics used in the literature to date. Our study provides a set of objective linguistic measures that can serve as metrics that further assist clinicians and policy makers in the mental health domain. 
The contributions of this paper are as follows:  
\begin{enumerate}
\item It provides a comprehensive set of cognitive and linguistics \textit{quantitative} metrics for schizophrenia patients language use;
\item We provide a translation of clinical observations of patient language use onto specific measurable linguistic cues that are mapped into advance NLP technology;
\item For the first time, our work leverages advances in the pragmatic NLP to measure patients' cognitive state (namely their levels of committed beliefs), personality traits, emotions, specificity and coherence;
\item We use LM with perplexity scores to measure both coherence and cohesion.

\end{enumerate}

\section{Related Work}
Language provides significant insight into the content of thought. It also reflects the presence  of impairments resulting from mental disorders such as schizophrenia. The predominent reflection of mental impairment for schizophrena is the lack of coherent text or speech. Accordingly, cohesion scores were first proposed as an indicator of predicting schizophrenia \citep{elvevaag2007quantifying} where they used Latent Semantic Analysis (LSA) as a feature extractor. This was further amplified by  \citep{bedi2015automated} where they measured the semantic coherence in disorganized speech captured by LSA, specifically where large amounts of language overlap was interpreted as coherent language. The study found that these features, together with syntactic markers of complexity, could predict later development of psychosis with 100\% accuracy using a convex hull algorithm. Later, \citet{corcoran2018prediction} used a logistic regression model to predict the onset of psychosis using coherence as measured by LSA combined with the usage of possessive pronouns. This approach showed an accuracy of 83\% in predicting the onset of psychosis with a cross-validation accuracy of 79\%. 

Metrics for Schizophrenia detection were investigated by  \citep{alqahtani2019understanding} where they used linguistic features such as referential cohesion, text ease, situation model, and readability in patients' and controls'  writing or speech to classify presence or absence of the disorder. The researchers trained Support  Vector  Machine  (SVM) and  Random  Forests  (RF)  models. The study  results showed that the situation model and readability performed the best among all cohesion features for the SVM  model yielding a 72\% F-score in the binary classification task of detecting whether a person (through their writing or speech) is a schizophrenia patient. 

Different from previous studies of schizophrenia, we propose measuring cohesion using language model perplexity. Moreover, we provide a comprehensive exploration of the language of patients relative to that of controls along the following linguistics cues:  coherence, emotion, personality, level of specificity, and level of committed belief.

\section{Data}
Our study comprises two datasets speech, \emph{LabSpeech}, and written text, \emph{LabWriting}. The data is obtained from schizophrenia patients and healthy controls. Both datasets are described in detail in \citep{kayi2018predictive}.\footnote{The authors of \cite{kayi2018predictive} kindly shared the data after we obtained IRB permission.} \emph{LabWriting} has $188$ participants who are native English-speakers between the ages of $18-50$ years, corresponding to 93 patients and 95 healthy controls. All participants are asked to write two paragraph-long essays: the first one is about their average Sunday and the second essay is about what makes them the angriest. The total number of writing samples collected from both patients and controls is $373$ pieces of text. 

The second dataset, \emph{LabSpeech}, includes three questions that prompt participants to describe some emotional and social events. Patients and controls are asked to describe (1) a picture, (2) their ideal day, and (3) their scariest experience. The total number of speech script samples collected from both patients and controls is $431$. Speech data is transcribed to text and a punctuation tool \citep{tilk2016} is used to add the missing punctuation.

\subsection{Superficial Descriptive statistics}
\tableref{tab:tab0} illustrates various descriptive statistics comparing and contrasting the  \emph{LabWriting} and \emph{LabSpeech} datasets. The results indicate  that healthy controls in both datasets are more verbose (produce more words and sentences) when answering questions  in both modalities, i.e. writing or speech. The mean values of the number of words and the number of sentences generated by Controls in \emph{LabWriting} are 141 and 7, respectively. However, the mean values are lower for Patients, with 110 and 6 for the same analysis. The Patients in \emph{LabSpeech} also score lower averages in all the descriptive features. These results are in line with a previous study \citep{de2020language} that individuals with schizophrenia speak less and use less complex sentences. * in \tableref{tab:tab0} indicates the higher results and statistically significant.

\begin{table}[t]
\begin{center}
\begin{tabular}{|l|cc|cc|}
\hline
\multicolumn{1}{|c|}{\multirow{2}{*}{\textbf{Descriptive}}} & \multicolumn{2}{c|}{\textbf{LabWriting}} & \multicolumn{2}{c|}{\textbf{LabSpeech}} \\ \cline{2-5} 
\multicolumn{1}{|c|}{} & P & C & P & C \\ \hline
Avg. \# words & 110 & \textbf{141*} & 220 & \textbf{277*} \\
Avg. \#sent. & 6 & \textbf{7*} & 11 & \textbf{14*} \\
sent./paragraph & 5.6 & \textbf{6.6*} & 11 & \textbf{14*} \\ \hline
\end{tabular}
\end{center}
\caption{Descriptive statistics for \emph{LabWriting} and \emph{LabSpeech} datasets. We present overall average number of words, overall average number of sentences and a finer grained average number of sentences per paragraph. \emph{P} denotes patient, and \emph{C} denotes control.}
\label{tab:tab0}
\end{table}

\section{Pragmatic Cues}
\subsection {Emotion}
Emotion refers to a person’s internal or external reaction to an event. This reaction can be expressed verbally, outwardly/visibly (e.g., frowning), or physiologically (e.g., crying) \citep{kring2010emotion}. Schizophrenia patients are often characterized as having disorganized thinking; however, according to \citep{kring2013emotion}, they still report their emotional experiences using the same general definitions of emotions (happy, sad, etc.) as persons who do not have schizophrenia. We use the EmoNet \footnote{\url{https://github.com/UBC-NLP/EmoNet}} \citep{abdul2017emonet} to obtain the eight core emotions (PL8), which are trust, anger, anticipation, disgust, joy, fear, sadness, and surprise. 

\subsection{Specificity}
Specificity in computational linguistic measures how much detail exists in a text \citep{louis2011automatic}. This is an important pragmatic concept and a characteristic of any text \citep{li2015fast}. We quantify this feature because schizophrenia may impacts one's language specificity. Hence, our hypothesis is that patients tend to write less specific paragraphs which lack references to any specific person, object, or event.   
We use \citep{ko2019domain} to measure a sentence specificity by indicating how many details exist in each sentence. This tool generates a rate for each sentence between 0 (general sentence) and 1 (detailed sentence). We also use Coh-Metrix \citep{graesser2004coh} to measure word hyponyms (i.e., word specificity) in a text. A higher value reflects an overall use of more specific words, which increases the ease and speed of text processing.

\subsection{Level of Committed belief (LCB)}
In natural language, the level of committed belief is a linguistic modality that indicates the author’s belief in a given proposition \citep{diab2009committed}. We measure this feature as it can detect an individual’s cognitive state. We want to explore this feature to test our hypothesis that patients with schizophrenia may hold strong beliefs towards their own propositions. We rely on a belief tagger \citep{rambow2016columbia} to label each sentence with the committed belief tags as \emph{(CB)} where someone \emph{(SW)} strongly believes in a proposition, Non-committed belief \emph{(NCB)} where SW reflects a weak belief in the proposition, and Non-Attributable Belief \emph{(NA)} where SW is not (or could not be) expressing a belief in the proposition (e.g., desires, questions, etc.). There is also the \emph{ROB} tag where SW’s intention is to report on someone else’s stated belief, regardless of whether or not they themselves believe it. The feature values are set to a binary 0 or 1 for each \emph{CB, NCB, NA, and ROB} corresponding to unseen or observed. The following text is an example from \emph{LabWriting}. 

\begin{quote} 
Every Sunday I usually \emph{$_{<cb-I>}$} get \emph{$_{</cb-I>}$} up and \emph{$_{<cb-I>}$} watch \emph{$_{</cb-I>}$} gospel shows on TV. I \emph{$_{<cb-I>}$} do \emph{$_{</cb-I>}$} my house chores and then \emph{$_{<cb-I>}$} watch \emph{$_{</cb-I>}$} other things on TV. Then later on I \emph{$_{<cb-I>}$} go \emph{$_{</cb-I>}$} down the street to the food resturants to \emph{$_{<na-I>}$} eat \emph{$_{</na-I>}$} something to eat.\footnote{Typos are in the original text.}
\end{quote}

We calculate the LCB as:
\begin{quote} 
    \centering 

\textit{\textbf{LCB $<tag>$} = total $<tag>$ in a text \textbf{/}  all LCB tags in the same text}
\end{quote} 

where $<tag>$ is one of the 4 LCB features: CB, NCB, NA, or ROB.

\subsection{Personality} In psychology, personality is the distinctive sets of behaviors, cognitions, and emotional patterns that derive from biological and environmental influence \citep{major2000encyclopedia}. We study the personality of patient and healthy controls in our datasets based on the famous Big-Five \citep{digman1990personality} personality measure, which are the following five traits: Extraversion (EXT), Neuroticism (NEU), Agreeableness (AGR), Conscientiousness (CON), and Openness (OPN). Neuroticism is characterized by a proclivity for negative emotions \citep{bono2007personality}. Individuals with high scores for neuroticism experience feelings such as anxiety, worry, fear, anger, frustration, depressed mood, and loneliness \citep{widiger2009neuroticism}. Extraversion indicates how outgoing and social a person is \citep{smelser2001international}. A low score in extraversion means an individual prefers to stay alone. We explore personality to test our hypothesis that patients with schizophrenia are high in neuroticism (emotionally unstable), especially if delusional, and low in extraversion \cite{horan2008affective}. We use \citep{kazameini2020personality} to predict personality traits for each text in our datasets. The model makes binary predictions of the author’s personality. 

\section{Cohesion Linguistic Features}

\subsection{Information Structure (Givenness)}
Latent Semantic Analysis (LSA) measures the semantic similarity/overlap between sentences or between paragraphs \citep{dennis2003introduction}. We use LSA to evaluate givenness, which is an information structure defined as a phenomenon where a speaker presumes that the listener is already familiar with the context of a discussion topic \citep{fery2016oxford}. The sentence is considered to be coherent when the average givenness score is high \citep{graesser2004coh}.

\subsection{Lexical Diversity}
Lexical diversity of a text is a measure of unique words (types), and consequently a measurement of different words that appear in the text compared to the total number of words (tokens) in that text  \citep{duran2004developmental} \cite {johansson2008lexical}. Type-token ratio (TTR), i.e., the ratio of types to tokens, is the most basic metric of lexical diversity \citep{duran2004developmental}.  When the number of types equals that of tokens in a text, all words are different, with TTR being equal to 1, and the lexical diversity of the text reaches its maximum possible value. Such a text, i.e., one with very high lexical diversity, is likely to be either low in cohesion because cohesion requires repetition of words or very short in length. After all, a naturally occurring longer text implies a greater frequency of the same word \citep{graesser2004coh}.

\subsection {Connectives}
The use of connecting words creates cohesive links between ideas and clauses and provides clues about text organization \citep{graesser2004coh}. We evaluate two types of connectives which are logic and temporal. The logic connectives are used to connect two or more ideas (such as \textit{and, or}). In contrast, temporal connectives are words or phrases that are used to indicate when something is taking place (such as \textit{first, until}).

\subsection{Syntactic Complexity}
Syntax refers to the arrangements of words and morphemes in forming larger units, such as phrases and clauses, ultimately resulting in well-formed sentences in a language \citep{crowhurst1983syntactic}. A tree-like structure, a syntactic tree, can visualize the arrangement of words in a sentence. A tree can be simple: containing basic structure like actor-action-object; or  complex, larger in size, with significant number of branches, and a complicated relationship among its different parts \citep{graesser2004coh}.

\subsection{Word Information}
All words in a sentence can be categorized as one of two types: a) Content words, such as nouns, verbs, adjectives, and adverbs, which primarily carry the semantic substance of the sentence and contribute to its meaning; and, b) Function words, such as prepositions, determiners, and pronouns, which primarily express the grammatical relationships among  content words without significant semantic content \citep{wilks1998d}.\footnote{We contend that this view is controversial since function words are critical to the meaning of utterances, however we would like to emphasize the qualitative difference between content words and function words.} Word Information refers to the notion that each word can be assigned a syntactic part-of-speech category and, with this assignment, be further rendered as a content or a function word, thus carrying either substantive or ``inconsequential" meaning \citep{graesser2004coh}.

\subsection{Language Model (LM) }
A Language model (LM) is the probability distribution over text \citep{bengio2003neural}. To analyze coherence in free text, we propose an approach based on LMs. We use a python library \emph{LM-scorer} \citep{Simone2020} to calculate probabilities of each word in a text and score sentences. The library uses the GPT2 model \citep{radford2019language} internally to provide a probability score for each next word.The sentence score (probability) is computed as the mean of tokens' probabilities.
For a given sentence, the LM predicts a higher score for a sentence that is more grammatically correct. Performance of LMs is commensurate with word information, content words tend to have lower probabilities compared to function words. 

We calculate multiple LM scores: the perplexity scores at sentence and paragraph level. Moreovere, we analyze the LM probabilities (scores) across two segmentation/levels: paragraph level and sentence level. We compare the performance of both levels using the means of statistical hypothesis testing. 

\subsubsection{\textbf{Analysis at Paragraph level}}
\begin{enumerate}

\item \textbf{Mean Sentence Probability:} For a given sentence, the LM predicts a higher score/probability for a sentence that is more grammatically and logically sound. We calculate the mean sentences probability in a text for each observation in each group (control/patient).

\item \textbf{Median Sentence Probability:} 
This statistic is calculated by taking the median of the probabilities of sentences. The justification for using this score is that the median, compared to the mean, is more robust to outliers. 
\end{enumerate}
\subsubsection{\textbf{Analysis at Sentence level}}
\begin{enumerate}
    \item {\textbf{Sentence probabilities:}} This statistic is extracted by aggregating LM individual sentence scores. Sentences scores for all patients and all controls are compared. The number of sentence probability scores analyzed is equivalent to the number of all sentences in the sample.
    \item{\textbf{Mean of the deltas in sentence probabilities:}} By using the sentences scores, the changes between the consecutive probability scores of the sentences in the paragraphs are extracted (deltas), and their average is calculated. The total number of this statistic is equivalent to the number of instances in the dataset. Our aim here is to check if the patient group has more fluctuations in their sentence probabilities.We use the following formula to calculate this statistic:
 \begin{multline*}[(log(s_1)-log(s_2)),(log(s_2)-log(s_3)),\\.....(log(s_{n-1})-log(s_n))]
 \end{multline*}
 \item {\textbf{Minimum deltas in sentence probabilities:}} 
    The minimum of changes in the sentence probabilities of consecutive sentences in each paragraph is calculated and compared. The total number of this statistic equals the number of instances in the dataset.
   \item {\textbf{Maximum of deltas in sentence probabilities:}} 
    Similar to the last statistic, the maximum of changes in the sentence probabilities of consecutive sentences in each paragraph are calculated and compared. The total number of this statistic equals the number of instances in the dataset. 
    \end{enumerate}

\section{Discussion of the Results}
\tableref{tab:tab00} and \tableref{tab:tab_11} illustrates the results of emotion analysis and specificity, respectively. \tableref{tab:tab_22} reports LCB averages and \tableref{tab:tab_33} summarizes personality percentages.    
\tableref{tab:tab_66} summarize the values of the cohesion linguistic features: Information Structure (Givenness), Connectives, Lexical Diversity, Syntactic Complexity, Syntactic Pattern Density, and Word Information. \tableref{tab:tab_44} and \tableref{tab:tab_55} show the values of the language model and perplexity scores. For each comparison criteria we compare the $p$-value to a significance level $\alpha$ = \textbf{0.05} to make conclusions about our hypotheses. \textbf{*} is used to indicate the results with a statistically significant $p$-value.
\begin{enumerate}

\item \textbf{Descriptive features} The $p$-values of the total number of sentences in both datasets are significant. As it is evident in the \figureref{fig1} and \figureref{fig2}, there is a noticeable difference between the distribution of this statistic between the two groups and it shows that Controls, on average, generate more sentences. 

\begin{figure}[ht]
   \centering
       \includegraphics[width=.45\textwidth]{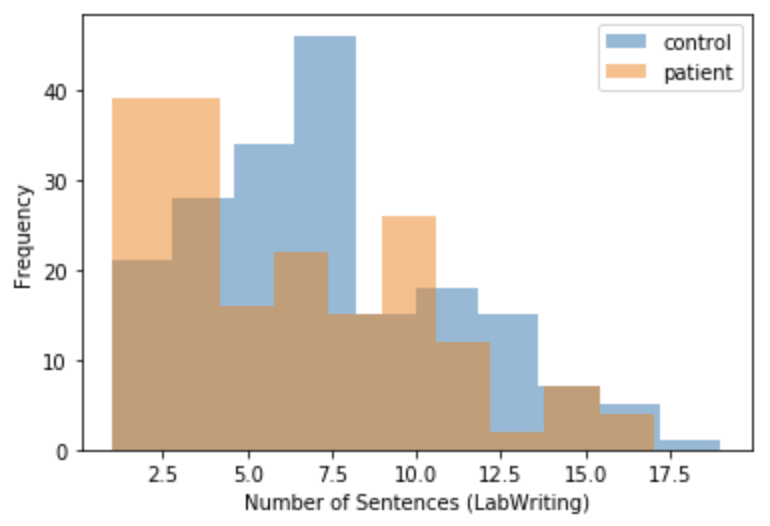}
 \caption{Text distribution in \emph{LabWriting}}
 \label{fig1}
\end{figure}

\begin{figure}[ht]
   \centering
       \includegraphics[width=.45\textwidth]{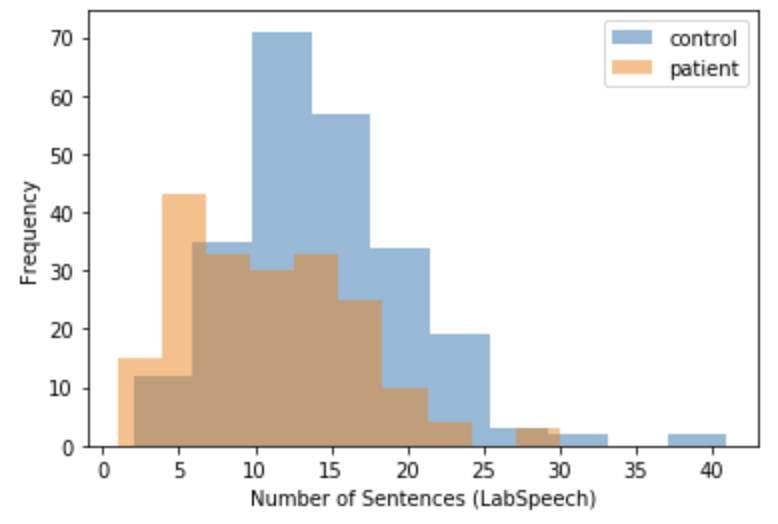}
 \caption{Text distribution in \emph{LabSpeech}}
 \label{fig2}
\end{figure}

\item \textbf{Emotion}
We hypothesise that Patients score high in fear. Our results show that Patients in both \emph{LabWriting} and \emph{LabSpeech} score high in fear ($p$--value = \textbf{0.002}) and ($p$-value=\textbf{0.004}), respectively. This result is consistent with a previous study \citep{suslow2003experience} which states that Patients tend to feel fear more often. Patients in \emph{LabWriting} score high in trust, and this may be due to interviewing them in a trustful environment.

\begin{table}[t]
\begin{center}
\begin{tabular}{|l|cc|cc|}
\hline
\multicolumn{1}{|c|}{\multirow{2}{*}{\textbf{Emotion}}} & \multicolumn{2}{c|}{\textbf{LabWriting}} & \multicolumn{2}{c|}{\textbf{LabSpeech}} \\ \cline{2-5} 
\multicolumn{1}{|c|}{} & P & C & P & C \\ \hline
\textbf{Anger}&0.159&0.162&\textbf{0.104*}&0.095\\
\textbf{Anticip.}&0.043&0.039&0.035&0.039\\
\textbf{Disgust}&0.086&0.093&0.117&0.112\\
\textbf{Fear}&\textbf{0.117*}&0.103&\textbf{0.181*}&0.167\\
\textbf{Joy}&0.372&0.381&0.297&0.311\\
\textbf{Sadness}&0.127&0.127&0.139&0.138\\
\textbf{Surprise}&0.077&0.079&0.117&0.127\\
\textbf{Trust}&\textbf{0.019*}&0.015&0.010&0.010\\\hline
\end{tabular}
\end{center}
\caption{Emotion results. The bold values above indicate the high means, and * indicates only the statistically significant values.}
\label{tab:tab00}
\end{table}

\item \textbf{Specificity} We hypothesise that Patients write less specific paragraphs. In the score of word hyponyms (Noun) as a measure of specificity, our results show that the Controls score significantly \textbf{higher} in \emph{LabWriting} ($p$-value  = \textbf{0.03}). Furthermore, Controls score higher in \emph{LabSpeech}, though not significantly. Specificity at sentence level is also significantly \textbf{higher}  in \emph{LabWriting} for Controls ($p$-value = \textbf{0.009}). However, there is no difference between Controls and Patients in \emph{LabSpeech}. It should be noted that the speech data are faithfully transcribed where pauses and filler words such as \emph{um, er, uh} can lower the quality of the speech relative the specificity model which is trained on native textual input hence making it challenging to capture specificity.

\begin{table}[t]
\begin{center}
\begin{tabular}{|l|cc|cc|}
\hline
\multicolumn{1}{|c|}{\multirow{2}{*}{\textbf{Specificity}}} & \multicolumn{2}{c|}{\textbf{LabWriting}} & \multicolumn{2}{c|}{\textbf{LabSpeech}} \\ \cline{2-5} 
\multicolumn{1}{|c|}{} & P & C & P & C \\ \hline
\textbf{Sent. level} &0.47&\textbf{0.48*}&0.39&0.39\\
\textbf{Hyponym}&5.85&\textbf{6.06*}&6.36&\textbf{6.45}\\\hline

\end{tabular}
\end{center}
\caption{Specificity results. The above table shows the average specificity at sentence level as well as word hyponyms (Noun). }
\label{tab:tab_11}
\end{table}

\item \textbf{LCB}
The hypothesis of this study states that Patients show more commitment to their beliefs.
Table~\ref{tab:tab_22} shows the results of LCB. It can be noticed that Patients in both datasets score \textbf{higher} in committed belief (CB) and Controls score \textbf{higher} in Non-committed belief (NCB). It confirms our hypothesis, and these findings coincide with a previous study \cite{kayi2018predictive} that patients with schizophrenia may show more commitment of their belief to propositions expressed in either modality, writing or speech. 

\begin{table}[t]
\begin{center}
\begin{tabular}{|l|cc|cc|}
\hline
\multicolumn{1}{|c|}{\multirow{2}{*}{\textbf{LCB}}} & \multicolumn{2}{c|}{\textbf{LabWriting}} & \multicolumn{2}{c|}{\textbf{LabSpeech}} \\ \cline{2-5} 
\multicolumn{1}{|c|}{} & P & C & P & C \\ \hline
\textbf{CB}&\textbf{0.52}&0.51&\textbf{0.60}&0.58\\
\textbf{NCB}&0.013&\textbf{0.020*}&0.04&\textbf{0.05}\\
\textbf{NA}&0.45&0.46&0.34&0.35\\
\textbf{ROB}&0.008&0.010&0.010&0.012\\\hline

\end{tabular}
\end{center}
\caption{LCB results.}
\label{tab:tab_22}
\end{table}

\item \textbf{Personality}
The hypothesis of this study states that Patients score high levels of neuroticism and low levels of extraversion. Table~\ref{tab:tab_33} reports the results of personality analysis. The results show that Patients in both datasets score \textbf{lower} in extroversion (EXT) ($p$-value = \textbf{0.03}) in \emph{LabWriting} and score \textbf{higher} in neuroticism (NEU) ($p$-value = \textbf{0.04}) in \emph{LabWriting}. These results are in line with previous studies \citep{camisa2005personality},\citep{horan2008affective}, \citep{smeland2017identification} which show that schizophrenia is associated with high levels of neuroticism and low levels of extraversion. We report all other personality traits in table~\ref{tab:tab_33}; However, our analysis mainly focuses on neuroticism and extraversion. Figure~\ref{per1} and Figure~\ref{per2} illustrate the personality results.

\begin{table}[t]
\begin{center}
\begin{tabular}{|l|cc|cc|}
\hline
\multicolumn{1}{|c|}{\multirow{2}{*}{\textbf{Personality}}} & \multicolumn{2}{c|}{\textbf{LabWriting}} & \multicolumn{2}{c|}{\textbf{LabSpeech}} \\ \cline{2-5} 
\multicolumn{1}{|c|}{} & P & C & P & C \\ \hline
\textbf{EXT}&34\%&\textbf{50\%*}&27\%&\textbf{30\%}\\
\textbf{NEU}&\textbf{52\%*}&40\%&\textbf{35\%}&27\%\\
\textbf{AGR}&60\%&\textbf{63\%}&81\%&\textbf{85\%}\\
\textbf{CON}&46\%&\textbf{54\%}&6.6\%&\textbf{7.3\%}\\
\textbf{OPN}&45\%&40\%&84\%&75\%\\\hline

\end{tabular}
\end{center}
\caption{Frequency Distribution of Personality Traits.}
\label{tab:tab_33}
\end{table}

\begin{figure}[ht]
   \centering
       \includegraphics[width=.45\textwidth]{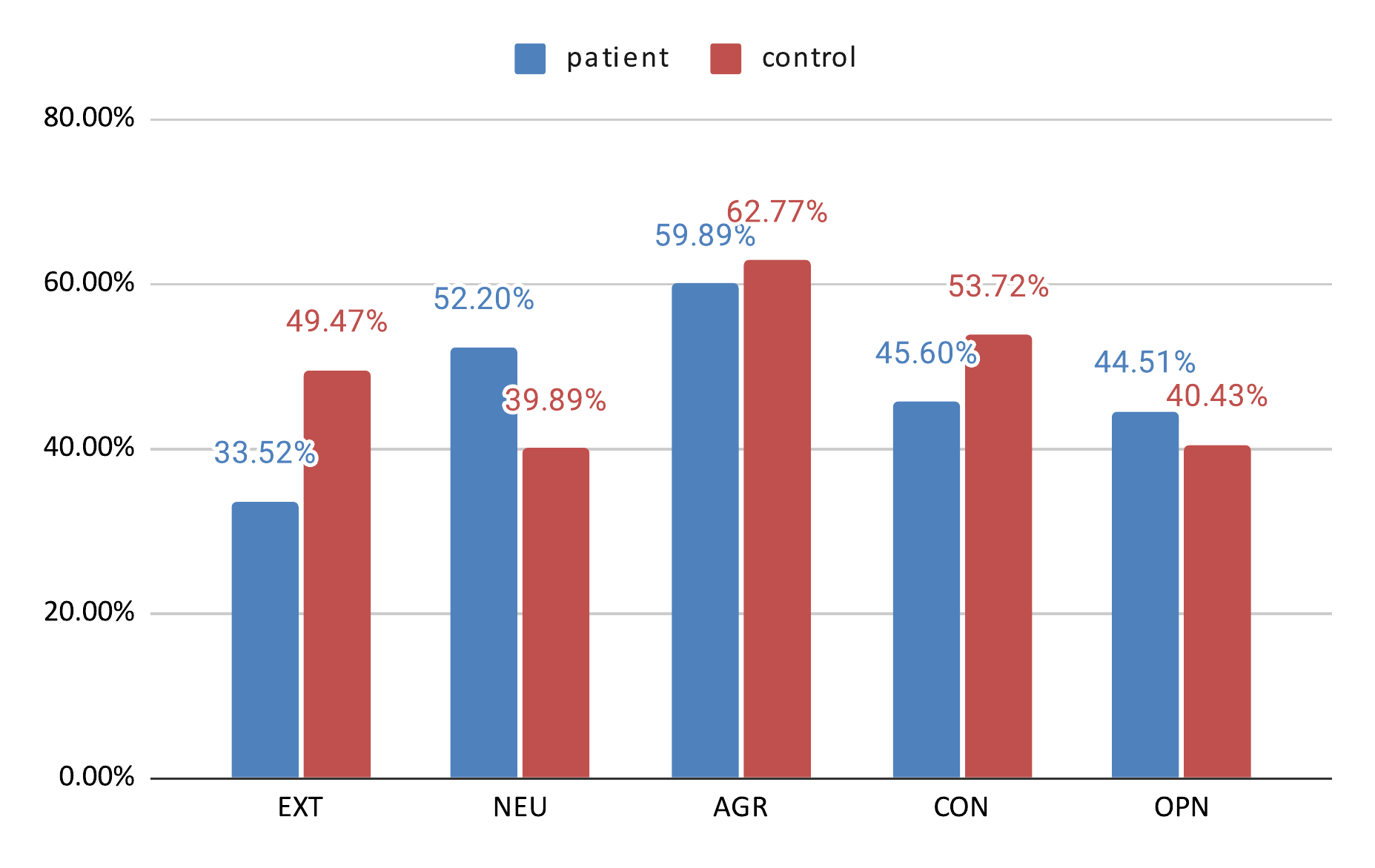}
 \caption{Big-5 personality traits of text in \emph{LabWriting}}
 \label{per1}
\end{figure}

\begin{figure}[ht]
   \centering
       \includegraphics[width=.45\textwidth]{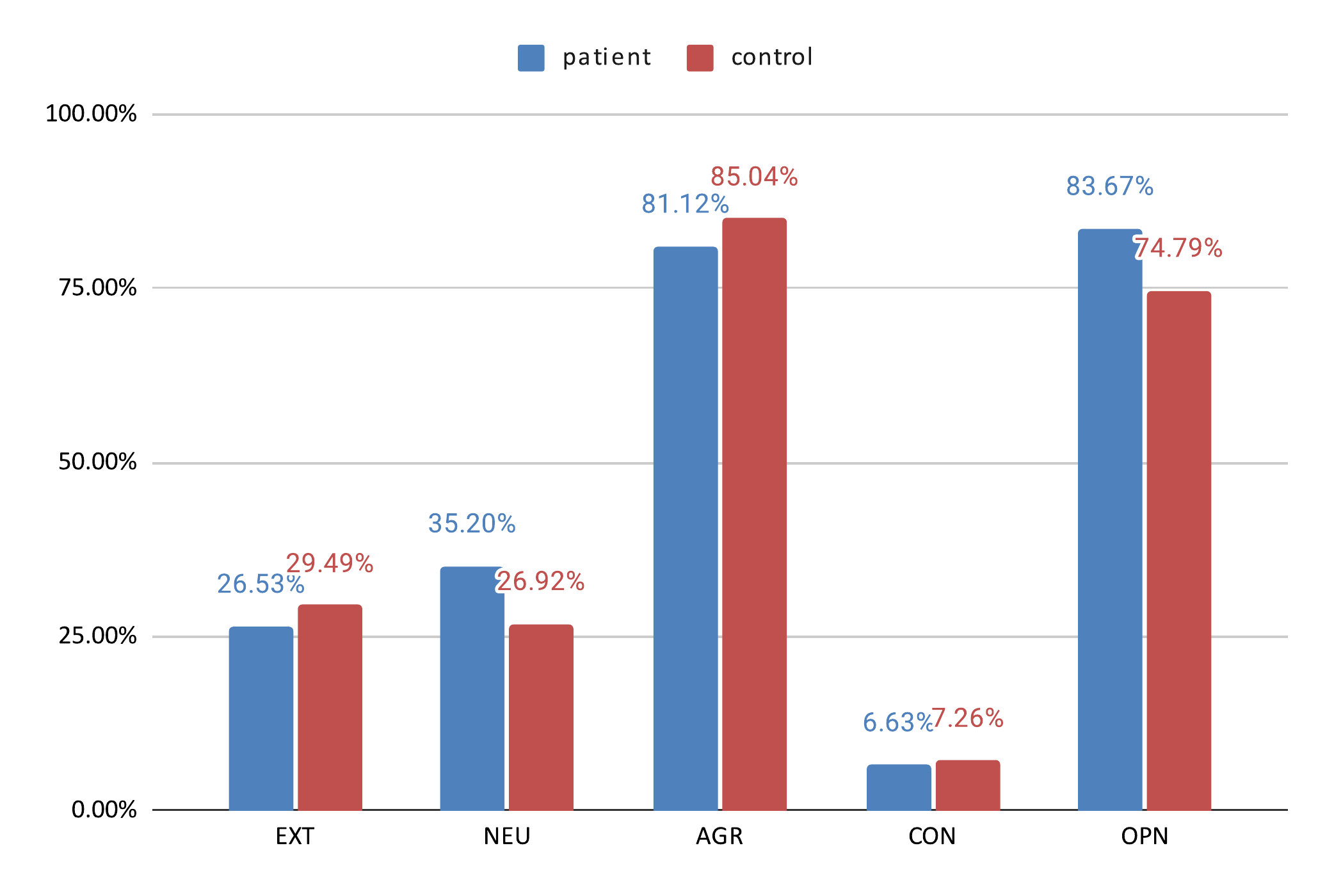}
 \caption{Big-5 personality traits of text in \emph{LabSpeech}}
 \label{per2}
\end{figure}

\item  \textbf{Information Structure (Givenness)} The average givenness per sentence of the schizophrenia patients is statistically significantly \textbf{lower} than that of the Controls in both \emph{LabWriting} ($p$-value =\textbf{0.001}) and \emph{LabSpeech} ($p$-value = \textbf{0.01}). Patients demonstrate challenges in recognizing things that others would find obvious and consequently question or repeat those. In addition, they present something that they have already mentioned earlier as completely new, compromising givenness.

\item \textbf{Lexical Diversity}
In the metric Type-token ratio (TTR) for all words, Patients scored \textbf{higher} than Controls, with the difference being statistically significant in both \emph{LabWriting} ($p$-value = \textbf {0.004}) and \emph{LabSpeech} ($p$-value = \textbf{0.0001}). The higher proportion of types by Patients stems from the fact that they produce more incomplete, indistinct, inaudible, or incomprehensible words or sounds and shorter sentences and utterances, struggling to reorganize their thoughts \citep{hinzen2019disturbing} \citep{merrill2017evidence}. These non-words, particularly shorter sentences, contribute to the higher TTRs for Patients.

Schizophrenic patients, however, are known to repeat words and phrases \citep{manschreck1985repetition}, and hence a basic TTR in itself is not a reliable indicator for distinguishing between Controls and Patients. TTR is only possible to apply when text or speech are of equal length. We thus compute two more metrics of lexical diversity, namely measure of textual lexical diversity (MTLD) and measure D vocabulary diversity (VocD), which allow comparison of lexical diversity of texts of unequal lengths. By these measures, we find text and speech of Controls to be lexically much more diverse, with $p$-values in the order of \textbf{${10^{-4}}$}. 

\item \textbf{Connectives} 
In the uses of logical, temporal, and extended temporal connectives in text and speech, Controls consistently score \textbf{higher}. The difference in scores is statistically significant in all three cases of speech which are logic, temporal, and extended temporal connectives 
with $p$-values respectively \textbf{0.03}, \textbf{0.04}, and \textbf{0.03}. In \emph{LabWriting}, the difference is, however, found to be statistically significant ($p$-value = \textbf{0.03}) only in the case of logical connectives. Our findings validate one of the decisive signs of schizophrenia, deficits of logical reasoning among patients \cite {willits2018evidence} \citep{mackinley2021linguistic}.

\item \textbf{Syntactic Complexity}
In addition to phonetic anomalies in terms of more pauses, loss of prosody, and mumbled sounds, syntactic and semantic conventions that govern the formation of sentences and ultimately the language are routinely violated by schizophrenia patients \citep{stein1993vocal}.
One of the manifestations of these violations is the decrease in the syntactic complexity of their writing and speech, resulting in disorganized language with poor content. According to all our three measures of syntactic complexity – SYNMEDpos, SYNMEDwrd, and SYNMEDlem – Controls demonstrate much \textbf{higher} syntactically complex text, with statistically significant differences from Patients  in all cases, except in \emph{LabSpeech}, in which the difference is nevertheless nearly significant. These results concur with previous studies \cite {kayi2018predictive} \citep{hinzen2019disturbing} which showed that a patient with schizophrenia alters the patterns of linguistic organization, which leads to increased syntactic errors. 
\item \textbf{Word Information}
In the usage of pronouns, our results show that Patients use the first-person pronouns, e.g., I, my, me, comparatively more, while Controls prefer first person plural, second-person, and third-person more. The differences are statistically significant in  text but not in speech. This result is in line with the previous study \citep{kayi2018predictive} \citep{tang2021natural}. One metric in which Controls score significantly \textbf{higher} in both \emph{LabWriting} and \emph{LabSpeech} is the average minimum word frequency in sentences. With Controls producing significantly longer writings or speeches, a greater frequency of words is necessary to maintain coherence and a logical flow in the text.   	

\begin{table*}[t]
\begin{center}
\begin{tabular}{|l|cc|cc|}
\hline
\multicolumn{1}{|c|}{\multirow{2}{*}{\textbf{Cohesion Linguistic Featuress}}} & \multicolumn{2}{c|}{\textbf{LabWriting}} & \multicolumn{2}{c|}{\textbf{LabSpeech}} \\ \cline{2-5} 
\multicolumn{1}{|c|}{} & P & C & P & C \\ \hline

\textbf{1. LSA}&&&&\\
Avg. givenness of each sentence&0.20&\textbf{0.23*}&0.31& \textbf{0.32*}\\
\hline
\textbf{2. Lexical Diversity}&&&&\\
Type token ratio (TTR) for all words&\textbf{0.63*}&0.60& \textbf{0.46*}&0.43\\
MTLD lexcical diversity measure for all words& 59.9&\textbf{68.82*}& 40.10&\textbf{44.95*}\\
VOC lexical diversity measure for all words&40.6&\textbf{67.7*}& 42.73&\textbf{52.20*}\\
\hline
\textbf{3. Connectives}&&&&\\
Score of logic connectives&47.1&\textbf{53.1*}&33.66&\textbf{38.31*}\\
Score of temporal connectives&24.5&\textbf{26.7}&12.24 &\textbf{14.74*}\\
Score of extended temporal connectives&24.3&\textbf{27.2}& 14.44&\textbf{18.04*}\\
\hline
\textbf{4. Syntactic Complexity}&&&&\\
SYNMEDpos*&0.56&\textbf{0.61*}&0.66&\textbf{0.68}\\
SYNMEDwrd*&0.73&\textbf{0.81*}&0.84&\textbf{0.87*}\\
SYNMEDlem*&0.71&\textbf{0.79*}&0.82&\textbf{0.84*}\\
\hline
\textbf{5. Word Information}&&&&\\
Score of pronouns, first person, single form&\textbf{96.6*}&86.11&\textbf{56.68}&55.03\\
Score of pronouns, first person, plural form&6.3&\textbf{10.2*}&5.24&\textbf{7.26}\\
Score of pronouns, second person&3.37&\textbf{6.22*}&\textbf{7.99}&7.33\\
Score of pronouns, third person, plural form&7.90&\textbf{12.25*}&7.20&\textbf{8.65}\\
Avg. minimum word frequency in sentences&0.83&\textbf{1.01*}&1.30&\textbf{1.45*}\\
\hline
\end{tabular}
\end{center}
\begin{tablenotes} 
\small 
\item \textbf{SYNMEDpos*}: mean minimum editorial distance score between adjacent sentences computed from POS. 
\item \textbf{SYNMEDwrd*}: minimum editorial distance score between adjacent sentences computed from words.  
\item \textbf{SYNMEDlem*}: This is the minimum editorial distance score between adjacent sentences from lemmas. 
\end{tablenotes} 
\caption{Coh-Metrix Linguistic Features Results}
\label{tab:tab_66}
\end{table*}

\item  \textbf{Language Model Analysis at Paragraph-Level} We measure the mean of the probabilities of the sentences and the corresponding medians to account for  outlier effects. Since both Patients and Controls produced an appreciable number of tokens per sentence, we find these probabilities lower for both groups. We are primarily interested in the comparison of the probabilities and find that the mean and median probabilities are significantly \textbf{lower} for Patients than for Controls in \emph{LabWriting}, with mean $p$-values of \textbf{0.01} and median $p$-values of \textbf{0.02}. The findings are in line with previous studies \citep{kuperberg2010language}, \citep{hinzen2015linguistics},\citep{de2020language} that schizophrenia patients often produce idiosyncratic expressions and hence less probable naturally occurring sentences. 

While the probabilities in \emph{LabSpeech} are \textbf{lower} for Patients, the differences in corresponding probabilities are not statistically significant at mean $p$-values of \textbf{0.524} and median $p$-values of \textbf{0.237}. This can be explained by the fact that Controls can exploit the time during writing better to their advantage to produce more organized and coherent text. Speech, on the other hand, is swift and spontaneous.

\item  \textbf{Language Model Analysis at Sentence Level}

In line with the mean and median of the probabilities of the sentences at the paragraph level, we compute the average of the probabilities of all sentences. This metric, average sentence probabilities, is also significantly \textbf{lower} for Patients (\textbf{0.109}) than for Controls (\textbf{0.117}) with ($p$-value=\textbf{0.0007}). The difference in \emph{LabSpeech} dataset, like that in the paragraph level, is again not statistically significant at ($p$-value=\textbf{0.175}).

The mean of changes in the sentence probabilities, computed to evaluate how strongly the sentence probabilities change from one sentence to another in a paragraph and consequently how much the sentences deviate from a coherent and logical flow, is \textbf{higher} for Controls  ($p$-value= \textbf{0.05}) in \emph{LabWriting}. Two other metrics related to this, the minimum and the maximum of changes in sentence probabilities, provide mixed, hence inconclusive, results. These probabilities, therefore may not be consistent indicators for the fluctuations we expected.

\begin{table*}[t]
\begin{center}
\begin{tabular}{|l|cc|cc|}
\hline
\multicolumn{1}{|c|}{\multirow{2}{*}{\textbf{Cohesion Linguistic Features}}} & \multicolumn{2}{c|}{\textbf{LabWriting}} & \multicolumn{2}{c|}{\textbf{LabSpeech}} \\ \cline{2-5} 
\multicolumn{1}{|c|}{} & P & C & P & C \\ \hline

\hline
\textbf{1. Analysis at Paragraph level}&&&&\\

- Mean of probabilities of sentences&0.110&\textbf{0.119*}&0.106&\textbf{0.107}\\
- Median of probabilities of sentences&0.107&\textbf{0.117*}&0.104&\textbf{0.107}\\
\hline
\textbf{2. Analysis at Sentence level}&&&&\\
-Sentence Probabilities&0.109&\textbf{0.117*}&0.103&\textbf{0.106}\\
-Mean of changes in sentence probabilities&-0.106&\textbf{-0.045*}&\textbf{-0.060}&-0.062\\
-Minimum of changes in sentence probabilities&-1.283&\textbf{-1.036*}&\textbf{-1.835*}&-2.107\\
-Maximum of changes in sentence probabilities&\textbf{1.036}&0.986&1.572&\textbf{1.902*}\\

\hline
\end{tabular}
\end{center}
\caption{The language model scores (probabilities) across different segmentation (levels)}
\label{tab:tab_44}
\end{table*}

\item \textbf{Perplexity}
Table~\ref{tab:tab_55} shows the results of perplexity. We compute it at two levels: the sentence level and the paragraph level, to determine how predictable the language of Patients is compared to that of Controls. In \emph{LabWriting}, the model is more perplexed for Patients in both levels, and the difference between the two groups is highly significant ($p$-value =\textbf{0.01}) at the paragraph level while ($p$-value =\textbf{0.00005}) at the sentence level. However, the results are not significant for \emph{LabSpeech} for any of the two levels.  
\begin{table}[t]
\begin{center}
\begin{tabular}{|l|cc|cc|}
\hline
\multicolumn{1}{|c|}{\multirow{2}{*}{\textbf{Levels}}} & \multicolumn{2}{c|}{\textbf{LabWriting}} & \multicolumn{2}{c|}{\textbf{LabSpeech}} \\ \cline{2-5} 
\multicolumn{1}{|c|}{} & P & C & P & C \\ \hline
\textbf{Sentence}&1.12&\textbf{1.10*}&\textbf{1.11}
&1.12\\
\textbf{Paragraph}&203.9&\textbf{150.4*}&245.5&\textbf{230.1}
\\\hline

\end{tabular}
\end{center}
\caption{Perplexity across different segmentation (levels)}
\label{tab:tab_55}
\end{table}
\end{enumerate}

\section{Conclusion}
Patients with schizophrenia experience different symptoms, some of which involve problems with concentration and memory, which in return may lead to disorganization in speech or behavior. Therefore, diagnosing this disorder early and correctly is extremely important as it may help alleviate the adverse effects on patients.

Among the linguistic features of cohesion investigated in this study, we found that Patients' scores are lower, with significant $p$-values in information structure (givenness), lexical diversity except for Type-token ratio (TTR), connectives, and syntactic complexity in both datasets. Among the pragmatic cues, we found that Patients' score high in fear, and their personality is associated with elevated neuroticism. They also show more commitment to their beliefs, and their average specificity at sentence and word levels is lower than Controls.

In the future, we plan to expand our analysis to other related mental health disorders. We also plan to explore the pragmatically motivated linguistics features of schizophrenia in other languages. 

\section*{Institutional Review Board (IRB)}
The authors of \cite{kayi2018predictive} kindly shared the data after we obtained IRB permission.

\bibliography{jmlr-sample}

\begin{thebibliography}{48}
\providecommand{\natexlab}[1]{#1}
\providecommand{\url}[1]{\texttt{#1}}
\expandafter\ifx\csname urlstyle\endcsname\relax
  \providecommand{\doi}[1]{doi: #1}\else
  \providecommand{\doi}{doi: \begingroup \urlstyle{rm}\Url}\fi

\bibitem[Abdul-Mageed and Ungar(2017)]{abdul2017emonet}
Muhammad Abdul-Mageed and Lyle Ungar.
\newblock Emonet: Fine-grained emotion detection with gated recurrent neural
  networks.
\newblock In \emph{Proceedings of the 55th annual meeting of the association
  for computational linguistics (volume 1: Long papers)}, pages 718--728, 2017.

\bibitem[AlQahtani et~al.(2019)AlQahtani, Kayi, and
  Diab]{alqahtani2019understanding}
Amal AlQahtani, Efsun Kayi, and Mona Diab.
\newblock Understanding cohesion in writings and speech of schizophrenia
  patients.
\newblock In \emph{2019 18th IEEE International Conference On Machine Learning
  And Applications (ICMLA)}, pages 364--369. IEEE, 2019.

\bibitem[Bedi et~al.(2015)Bedi, Carrillo, Cecchi, Slezak, Sigman, Mota,
  Ribeiro, Javitt, Copelli, and Corcoran]{bedi2015automated}
Gillinder Bedi, Facundo Carrillo, Guillermo~A Cecchi, Diego~Fern{\'a}ndez
  Slezak, Mariano Sigman, Nat{\'a}lia~B Mota, Sidarta Ribeiro, Daniel~C Javitt,
  Mauro Copelli, and Cheryl~M Corcoran.
\newblock Automated analysis of free speech predicts psychosis onset in
  high-risk youths.
\newblock \emph{npj Schizophrenia}, 1\penalty0 (1):\penalty0 1--7, 2015.

\bibitem[Bengio et~al.(2003)Bengio, Ducharme, Vincent, and
  Jauvin]{bengio2003neural}
Yoshua Bengio, R{\'e}jean Ducharme, Pascal Vincent, and Christian Jauvin.
\newblock A neural probabilistic language model.
\newblock \emph{Journal of machine learning research}, 3\penalty0
  (Feb):\penalty0 1137--1155, 2003.

\bibitem[Bono and Vey(2007)]{bono2007personality}
Joyce~E Bono and Meredith~A Vey.
\newblock Personality and emotional performance: Extraversion, neuroticism, and
  self-monitoring.
\newblock \emph{Journal of occupational health psychology}, 12\penalty0
  (2):\penalty0 177, 2007.

\bibitem[Camisa et~al.(2005)Camisa, Bockbrader, Lysaker, Rae, Brenner, and
  O'Donnell]{camisa2005personality}
Kathryn~M Camisa, Marcia~A Bockbrader, Paul Lysaker, Lauren~L Rae, Colleen~A
  Brenner, and Brian~F O'Donnell.
\newblock Personality traits in schizophrenia and related personality
  disorders.
\newblock \emph{Psychiatry research}, 133\penalty0 (1):\penalty0 23--33, 2005.

\bibitem[Corcoran et~al.(2018)Corcoran, Carrillo, Fern{\'a}ndez-Slezak, Bedi,
  Klim, Javitt, Bearden, and Cecchi]{corcoran2018prediction}
Cheryl~M Corcoran, Facundo Carrillo, Diego Fern{\'a}ndez-Slezak, Gillinder
  Bedi, Casimir Klim, Daniel~C Javitt, Carrie~E Bearden, and Guillermo~A
  Cecchi.
\newblock Prediction of psychosis across protocols and risk cohorts using
  automated language analysis.
\newblock \emph{World Psychiatry}, 17\penalty0 (1):\penalty0 67--75, 2018.

\bibitem[Crowhurst(1983)]{crowhurst1983syntactic}
Marion Crowhurst.
\newblock Syntactic complexity and writing quality: A review.
\newblock \emph{Canadian Journal of Education/Revue canadienne de l'education},
  pages 1--16, 1983.

\bibitem[De~Boer et~al.(2020)De~Boer, van Hoogdalem, Mandl, Brummelman, Voppel,
  Begemann, van Dellen, Wijnen, and Sommer]{de2020language}
JN~De~Boer, M~van Hoogdalem, RCW Mandl, J~Brummelman, AE~Voppel, MJH Begemann,
  E~van Dellen, FNK Wijnen, and IEC Sommer.
\newblock Language in schizophrenia: relation with diagnosis, symptomatology
  and white matter tracts.
\newblock \emph{npj Schizophrenia}, 6\penalty0 (1):\penalty0 1--10, 2020.

\bibitem[Dennis et~al.(2003)Dennis, Landauer, Kintsch, and
  Quesada]{dennis2003introduction}
Simon Dennis, Tom Landauer, Walter Kintsch, and Jose Quesada.
\newblock Introduction to latent semantic analysis.
\newblock In \emph{25th Annual Meeting of the Cognitive Science Society.
  Boston, Mass}, page~25, 2003.

\bibitem[Diab et~al.(2009)Diab, Levin, Mitamura, Rambow, Prabhakaran, and
  Guo]{diab2009committed}
Mona Diab, Lori Levin, Teruko Mitamura, Owen Rambow, Vinodkumar Prabhakaran,
  and Weiwei Guo.
\newblock Committed belief annotation and tagging.
\newblock In \emph{Proceedings of the Third Linguistic Annotation Workshop (LAW
  III)}, pages 68--73, 2009.

\bibitem[Digman(1990)]{digman1990personality}
John~M Digman.
\newblock Personality structure: Emergence of the five-factor model.
\newblock \emph{Annual review of psychology}, 41\penalty0 (1):\penalty0
  417--440, 1990.

\bibitem[Dur{\'a}n et~al.(2004)Dur{\'a}n, Malvern, Richards, and
  Chipere]{duran2004developmental}
Pilar Dur{\'a}n, David Malvern, Brian Richards, and Ngoni Chipere.
\newblock Developmental trends in lexical diversity.
\newblock \emph{Applied Linguistics}, 25\penalty0 (2):\penalty0 220--242, 2004.

\bibitem[Elvevag and Goldberg(2000)]{elvevag2000cognitive}
Brita Elvevag and Terry~E Goldberg.
\newblock Cognitive impairment in schizophrenia is the core of the disorder.
\newblock \emph{Critical Reviews™ in Neurobiology}, 14\penalty0 (1), 2000.

\bibitem[Elvev{\aa}g et~al.(2007)Elvev{\aa}g, Foltz, Weinberger, and
  Goldberg]{elvevaag2007quantifying}
Brita Elvev{\aa}g, Peter~W Foltz, Daniel~R Weinberger, and Terry~E Goldberg.
\newblock Quantifying incoherence in speech: an automated methodology and novel
  application to schizophrenia.
\newblock \emph{Schizophrenia research}, 93\penalty0 (1-3):\penalty0 304--316,
  2007.

\bibitem[F{\'e}ry and Ishihara(2016)]{fery2016oxford}
Caroline F{\'e}ry and Shinichiro Ishihara.
\newblock \emph{The Oxford handbook of information structure}.
\newblock Oxford University Press, 2016.

\bibitem[Gao et~al.(2019)Gao, Zhong, Preo\c{t}iuc-Pietro, and
  Li]{gao2019specificity}
Yifan Gao, Yang Zhong, Daniel Preo\c{t}iuc-Pietro, and Junyi~Jessy Li.
\newblock Predicting and analyzing language specificity in social media posts.
\newblock In \emph{Proceedings of AAAI}, 2019.

\bibitem[Graesser et~al.(2004)Graesser, McNamara, Louwerse, and
  Cai]{graesser2004coh}
Arthur~C Graesser, Danielle~S McNamara, Max~M Louwerse, and Zhiqiang Cai.
\newblock Coh-metrix: Analysis of text on cohesion and language.
\newblock \emph{Behavior research methods, instruments, \& computers},
  36\penalty0 (2):\penalty0 193--202, 2004.

\bibitem[Hinzen and Rossell{\'o}(2015)]{hinzen2015linguistics}
Wolfram Hinzen and Joana Rossell{\'o}.
\newblock The linguistics of schizophrenia: thought disturbance as language
  pathology across positive symptoms.
\newblock \emph{Frontiers in psych ology}, 6:\penalty0 971, 2015.

\bibitem[Hinzen et~al.(2019)Hinzen, {\c{C}}okal, Zimmerer, Turkington, Ferrier,
  Varley, and Watson]{hinzen2019disturbing}
Wolfram Hinzen, Derya {\c{C}}okal, Vitor~C Zimmerer, Douglas Turkington,
  I~Nicol Ferrier, Rosemary Varley, and Stuart Watson.
\newblock Disturbing the rhythm of thought: speech pausing patterns in
  schizophrenia, with and without formal thought disorder.
\newblock \emph{Plos One. 2019; 14 (5): e0217404. DOI: 10.1371/journal. pone.
  0217404}, 2019.

\bibitem[Horan et~al.(2008)Horan, Blanchard, Clark, and
  Green]{horan2008affective}
William~P Horan, Jack~J Blanchard, Lee~Anna Clark, and Michael~F Green.
\newblock Affective traits in schizophrenia and schizotypy.
\newblock \emph{Schizophrenia bulletin}, 34\penalty0 (5):\penalty0 856--874,
  2008.

\bibitem[Johansson(2008)]{johansson2008lexical}
Victoria Johansson.
\newblock Lexical diversity and lexical density in speech and writing: A
  developmental perspective.
\newblock \emph{Working papers/Lund University, Department of Linguistics and
  Phonetics}, 53:\penalty0 61--79, 2008.

\bibitem[Kayi et~al.(2018)Kayi, Diab, Pauselli, Compton, and
  Coppersmith]{kayi2018predictive}
Efsun~Sarioglu Kayi, Mona Diab, Luca Pauselli, Michael Compton, and Glen
  Coppersmith.
\newblock Predictive linguistic features of schizophrenia.
\newblock \emph{arXiv preprint arXiv:1810.09377}, 2018.

\bibitem[Kazameini et~al.(2020)Kazameini, Fatehi, Mehta, Eetemadi, and
  Cambria]{kazameini2020personality}
Amirmohammad Kazameini, Samin Fatehi, Yash Mehta, Sauleh Eetemadi, and Erik
  Cambria.
\newblock Personality trait detection using bagged svm over bert word embedding
  ensembles.
\newblock \emph{arXiv preprint arXiv:2010.01309}, 2020.

\bibitem[Kerns and Berenbaum(2002)]{kerns2002cognitive}
John~G Kerns and Howard Berenbaum.
\newblock Cognitive impairments associated with formal thought disorder in
  people with schizophrenia.
\newblock \emph{Journal of abnormal psychology}, 111\penalty0 (2):\penalty0
  211, 2002.

\bibitem[Ko et~al.(2019)Ko, Durrett, and Li]{ko2019domain}
Wei-Jen Ko, Greg Durrett, and Junyi~Jessy Li.
\newblock Domain agnostic real-valued specificity prediction.
\newblock In \emph{AAAI}, 2019.

\bibitem[Kring and Caponigro(2010)]{kring2010emotion}
Ann~M Kring and Janelle~M Caponigro.
\newblock Emotion in schizophrenia: where feeling meets thinking.
\newblock \emph{Current directions in psychological science}, 19\penalty0
  (4):\penalty0 255--259, 2010.

\bibitem[Kring and Elis(2013)]{kring2013emotion}
Ann~M Kring and Ori Elis.
\newblock Emotion deficits in people with schizophrenia.
\newblock \emph{Annual review of clinical psychology}, 9:\penalty0 409--433,
  2013.

\bibitem[Kuperberg(2010)]{kuperberg2010language}
Gina~R Kuperberg.
\newblock Language in schizophrenia part 1: an introduction.
\newblock \emph{Language and linguistics compass}, 4\penalty0 (8):\penalty0
  576--589, 2010.

\bibitem[Li and Nenkova(2015)]{li2015fast}
Junyi Li and Ani Nenkova.
\newblock Fast and accurate prediction of sentence specificity.
\newblock In \emph{AAAI}, 2015.

\bibitem[Louis and Nenkova(2011)]{louis2011automatic}
Annie Louis and Ani Nenkova.
\newblock Automatic identification of general and specific sentences by
  leveraging discourse annotations.
\newblock In \emph{Proceedings of 5th international joint conference on natural
  language processing}, pages 605--613, 2011.

\bibitem[Mackinley et~al.(2021)Mackinley, Chan, Ke, Dempster, and
  Palaniyappan]{mackinley2021linguistic}
Michael Mackinley, Jenny Chan, Hanna Ke, Kara Dempster, and Lena Palaniyappan.
\newblock Linguistic determinants of formal thought disorder in first episode
  psychosis.
\newblock \emph{Early intervention in psychiatry}, 15\penalty0 (2):\penalty0
  344--351, 2021.

\bibitem[Major et~al.(2000)Major, Cozzarelli, Horowitz, Colyer, Fuchs, Shapiro,
  Stoiber, Malt, Teo, Winter, et~al.]{major2000encyclopedia}
Brenda Major, Catherine Cozzarelli, Mardi~J Horowitz, Peter~J Colyer, Lynn~S
  Fuchs, Edward~S Shapiro, Karen~Callan Stoiber, Ulrik~Fredrik Malt, Thomas
  Teo, David~G Winter, et~al.
\newblock Encyclopedia of psychology: 8 volume set.
\newblock \emph{New York and Washington: Oxford University Press and the
  American Psychological Association}, 2000.

\bibitem[Manschreck et~al.(1985)Manschreck, Maher, Hoover, and
  Ames]{manschreck1985repetition}
Theo~C Manschreck, Brendan~A Maher, Toni~M Hoover, and Donna Ames.
\newblock Repetition in schizophrenic speech.
\newblock \emph{Language and Speech}, 28\penalty0 (3):\penalty0 255--268, 1985.

\bibitem[Merrill et~al.(2017)Merrill, Karcher, Cicero, Becker, Docherty, and
  Kerns]{merrill2017evidence}
Anne~M Merrill, Nicole~R Karcher, David~C Cicero, Theresa~M Becker, Anna~R
  Docherty, and John~G Kerns.
\newblock Evidence that communication impairment in schizophrenia is associated
  with generalized poor task performance.
\newblock \emph{Psychiatry research}, 249:\penalty0 172--179, 2017.

\bibitem[Radford et~al.(2019)Radford, Wu, Child, Luan, Amodei, Sutskever,
  et~al.]{radford2019language}
Alec Radford, Jeffrey Wu, Rewon Child, David Luan, Dario Amodei, Ilya
  Sutskever, et~al.
\newblock Language models are unsupervised multitask learners.
\newblock \emph{OpenAI blog}, 1\penalty0 (8):\penalty0 9, 2019.

\bibitem[Rambow et~al.(2016)Rambow, Yu, Radeva, Fabbri, Hidey, Peng, McKeown,
  Muresan, Hamidian, Diab, et~al.]{rambow2016columbia}
Owen Rambow, Tao Yu, Axinia Radeva, Alexander~R Fabbri, Christopher Hidey,
  Tianrui Peng, Kathleen~R McKeown, Smaranda Muresan, Sardar Hamidian, Mona~T
  Diab, et~al.
\newblock The columbia-gwu system at the 2016 tac kbp best evaluation.
\newblock In \emph{TAC}, 2016.

\bibitem[Seeber and Cadenhead(2005)]{seeber2005does}
Katherine Seeber and Kristin~S Cadenhead.
\newblock How does studying schizotypal personality disorder inform us about
  the prodrome of schizophrenia?
\newblock \emph{Current Psychiatry Reports}, 7\penalty0 (1):\penalty0 41--50,
  2005.

\bibitem[Simone(2020)]{Simone2020}
Primarosa Simone.
\newblock lm-scorer.
\newblock \url{https://github.com/simonepri/lm-scorer}, 2020.

\bibitem[Smeland et~al.(2017)Smeland, Wang, Lo, Li, Frei, Witoelar, Tesli,
  Hinds, Tung, Djurovic, et~al.]{smeland2017identification}
Olav~B Smeland, Yunpeng Wang, Min-Tzu Lo, Wen Li, Oleksandr Frei, Aree
  Witoelar, Martin Tesli, David~A Hinds, Joyce~Y Tung, Srdjan Djurovic, et~al.
\newblock Identification of genetic loci shared between schizophrenia and the
  big five personality traits.
\newblock \emph{Scientific reports}, 7\penalty0 (1):\penalty0 1--9, 2017.

\bibitem[Smelser et~al.(2001)Smelser, Baltes, et~al.]{smelser2001international}
Neil~J Smelser, Paul~B Baltes, et~al.
\newblock \emph{International encyclopedia of the social \& behavioral
  sciences}, volume~11.
\newblock Elsevier Amsterdam, 2001.

\bibitem[Stein(1993)]{stein1993vocal}
Johanna Stein.
\newblock Vocal alterations in schizophrenic speech.
\newblock \emph{Journal of Nervous and Mental Disease}, 1993.

\bibitem[Suslow et~al.(2003)Suslow, Roestel, Ohrmann, and
  Arolt]{suslow2003experience}
Thomas Suslow, Cornelia Roestel, Patricia Ohrmann, and Volker Arolt.
\newblock The experience of basic emotions in schizophrenia with and without
  affective negative symptoms.
\newblock \emph{Comprehensive psychiatry}, 44\penalty0 (4):\penalty0 303--310,
  2003.

\bibitem[Tang et~al.(2021)Tang, Kriz, Cho, Park, Harowitz, Gur, Bhati, Wolf,
  Sedoc, and Liberman]{tang2021natural}
Sunny~X Tang, Reno Kriz, Sunghye Cho, Suh~Jung Park, Jenna Harowitz, Raquel~E
  Gur, Mahendra~T Bhati, Daniel~H Wolf, Jo{\~a}o Sedoc, and Mark~Y Liberman.
\newblock Natural language processing methods are sensitive to sub-clinical
  linguistic differences in schizophrenia spectrum disorders.
\newblock \emph{npj Schizophrenia}, 7\penalty0 (1):\penalty0 1--8, 2021.

\bibitem[Tilk and Alum{\"a}e(2016)]{tilk2016}
Ottokar Tilk and Tanel Alum{\"a}e.
\newblock Bidirectional recurrent neural network with attention mechanism for
  punctuation restoration.
\newblock In \emph{Interspeech 2016}, 2016.

\bibitem[Widiger(2009)]{widiger2009neuroticism}
Thomas~A Widiger.
\newblock \emph{Neuroticism}, volume~11.
\newblock The Guilford Press, 2009.

\bibitem[Wilks(1998)]{wilks1998d}
Yorick Wilks.
\newblock D. arnold, l. balkan, r. lee humphries, s. meijer and l. sadler.
  machine translation: an introductory guide. ncc blackwell, oxford,
  1994.(hardback isbn 1-85554-246-3 $49.95/{\pounds} 40.00; paperback isbn
  1-85554-217-x $19.95/{\pounds} 18.99) viii+ 240 pages.
\newblock \emph{Natural Language Engineering}, 4\penalty0 (4):\penalty0
  363--382, 1998.

\bibitem[Willits et~al.(2018)Willits, Rubin, Jones, Minor, and
  Lysaker]{willits2018evidence}
Jon~A Willits, Timothy Rubin, Michael~N Jones, Kyle~S Minor, and Paul~H
  Lysaker.
\newblock Evidence of disturbances of deep levels of semantic cohesion within
  personal narratives in schizophrenia.
\newblock \emph{Schizophrenia Research}, 197:\penalty0 365--369, 2018.

\end{thebibliography}

\appendix



\end{document}